\pdfoutput=1

\documentclass[11pt]{article}

\usepackage[final]{acl}

\usepackage{times}
\usepackage{latexsym}

\usepackage[T1]{fontenc}

\usepackage[utf8]{inputenc}

\usepackage{microtype}

\usepackage{inconsolata}

\usepackage{graphicx}

\usepackage{bbm}
\usepackage{makecell}
\usepackage{multirow}

\usepackage{enumitem}
\usepackage{amsmath}
\newcommand\numberthis{\addtocounter{equation}{1}\tag{\theequation}}
\usepackage{amssymb}
\usepackage{amsthm}

\theoremstyle{definition}

\theoremstyle{remark}

\DeclareMathOperator{\E}{\mathbb{E}}

\title{Estimating Causal Effects of Text Interventions Leveraging LLMs}

\author{
 \textbf{Siyi Guo$^*$},
 \textbf{Myrl G. Marmarelis$^*$},
 \textbf{Fred Morstatter},
 \textbf{Kristina Lerman}
\\
 USC Information Sciences Institute \\
  Marina del Rey, California, USA \\
\\
 \small{
   \textbf{Correspondence:} \href{mailto:siyiguo@isi.edu}{siyiguo@isi.edu}
 }
}

\begin{document}

\maketitle

\def\thefootnote{*}\footnotetext{These authors contributed equally to this work.}\def\thefootnote{\arabic{footnote}}

\begin{abstract}
Quantifying the effects of textual interventions in social systems, such as reducing anger in social media posts to see its impact on engagement, is challenging. Real-world interventions are often infeasible, necessitating reliance on observational data. Traditional causal inference methods, typically designed for binary or discrete treatments, are inadequate for handling the complex, high-dimensional textual data. This paper addresses these challenges by proposing \textsc{CausalDANN}, a novel approach to estimate causal effects using text transformations facilitated by large language models (LLMs). Unlike existing methods, our approach accommodates arbitrary textual interventions and leverages text-level classifiers with domain adaptation ability to produce robust effect estimates against domain shifts, even when only the control group is observed. This flexibility in handling various text interventions is a key advancement in causal estimation for textual data, offering opportunities to better understand human behaviors and develop effective interventions within social systems.
\end{abstract}




\section{Introduction}\label{sec:intro}
Causal inference is essential for studying social phenomena from observational data, as it distinguishes true effects from spurious correlations~\cite{oktay2010causal}. Unlike predictive models, causal inference explores counterfactual scenarios, offering deeper insights into individual and social behaviors~\cite{adhikari2023inferring,russo2024stranger} and informing effective interventions~\cite{pan2016health,kleven2024family}. Given that much of online data is textual, estimating the causal effects of language presents an important challenge~\cite{russo23, egami22, weld22_nlp, gligoric19}. 
Without causal inference, studies risk hidden biases and misleading conclusions~\citep{leeb2025causality}. Given the complexity of human behavior and society, robust causal methods are crucial for producing reliable, actionable insights.

\begin{figure}[t]
    \centering
    \includegraphics[width=1\linewidth]{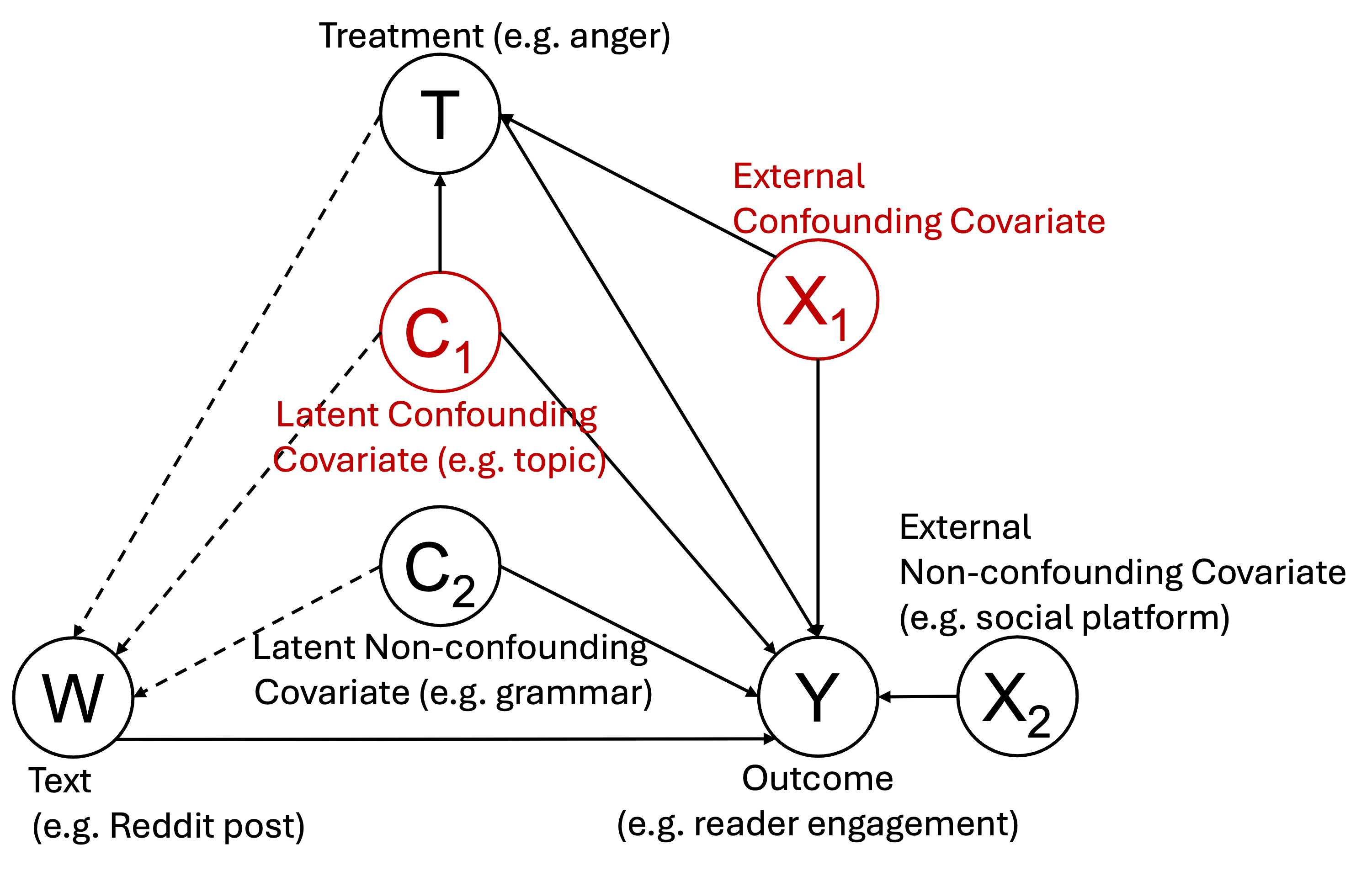}
    \caption{The causal diagram of the problem setup. We aim to estimate the effect from the treatment T to the outcome Y, accounting for confounding and/or non-confounding covariates. The dashed arrows represent latent relationships.}
    \label{fig:DAG}
\end{figure}

However, causal inference on utterances of language presents unique challenges both for identification and estimation~\citep{chen23_nlp,feder22}. Consider the question of whether angrier social media posts receive more attention (Figure~\ref{fig:DAG}). 
The key challenge is that the treatment variable ``anger,'' is latently embedded in text, never directly observed and must be inferred, which further complicates causal-effect estimation~\citep{pryzant21}. Any biases in anger estimation can distort the causal estimate. This is the fundamental challenge of identifying the causal effect of a treatment that is itself a latent attribute of text. Another challenge is that identifying and adjusting for confounders is particularly difficult in observational studies, making results susceptible to various biases from hidden confounding.


We propose \textbf{\textsc{CausalDANN}}, a framework that crafts a hypothetical intervention on the observed text, which can be implemented as a text transformation using a large language model (LLM), and estimates
the effect by predicting the outcomes for both observed and
transformed data. 
This approach enables causal inference via language even when an intervention group is missing, overcoming a key limitation of conventional methods.

To the best of our knowledge, we are the first to address the estimation of causal effects from direct interventions on text. Our first contribution is formulating the intervention space as text transformations. Such interventions can be implemented by prompting an LLM, like to rephrase text to express more anger while preserving all other properties. The treatment variable in our potential-outcomes framework~\citep{rubin74} is considered to be a whole text, 
allowing for causal analysis even in the absence of an intervention group. It also reduces the need for identifying and adjusting for confounders already embedded in text.

Our second contribution is a method to predict unobserved outcomes of text interventions. After transformation, we have both non-intervened (observed) and intervened text, but outcomes for the latter remain unobserved. Predicting the outcomes for unseen data is challenging due to domain shift between the observed training data and the unobserved intervened data~\cite{ryu2022knowledge,ma2025llm}. To tackle this, we propose \textsc{CausalDANN}, which adapts domain adaptation neural network (DANN)~\cite{ganin2015unsupervised} as the outcome predictor. Experiments show that DANN leads to less bias in causal estimation than alternatives like inverse propensity weighting (IPW) and doubly robust method (DR).

We evaluate \textsc{CausalDANN} on three semi-synthetic datasets constructed using LLM simulations, and find that it outperforms baselines for estimating causal effects of direct text interventions. 
In our evaluation, we analyze potential sources of bias, but recognize the larger need to evaluate and mitigate bias in LLM-generated text in future applications, which can impact whether the chosen LLM transformations truly isolate the causal variable while preserving all else. Future works may assess whether this framework generalizes to real-world settings beyond the semi-synthetic datasets.
\footnote{Our code can be found at \url{https://github.com/fionasguo/CausalDANN}.}

\section{Related Works} \label{sec:related_works}

\paragraph{Causal Inference with Text}
The field of causal inference in machine learning is rapidly advancing, as does its intersection with language modeling.
Prior works on causal effect estimation with text~\citep{sridhar22} does not usually consider the text as the treatment variable. Rather, text embeddings are used as covariate information~\citep{veitch20}, or some discrete coding of the text is used as the treatment~\citep{pryzant21, egami22, cheng2022estimating,zhou-he-2023-causal,jiang2024media}. 
As far as we know, we are the first to introduce a method for estimating the effects of direct interventions on text. 

Efforts to mitigate spurious correlations in text classifiers~\citep{veitch21, bansal23} mirrors some of the techniques used for causal effect estimation, as their statistical implications resemble confounding.
Generally, spurious correlations are of concern for model generalization~\citep{feder23}, and causal effect estimation can be seen as a special case of generalization to the interventional domain. 


\paragraph{LLMs for Causal Inference}
LLMs offer potential for advancing causal inference with text~\citep{kiciman23, zhang23, ashwani24, jin23_llm}. However, spurious correlations in large text corpora often obscure causal relationships~\citep{zevcevic23}. This weakness may stem from a fundamental limitation of LLM training—causal inference requires extrapolating to novel conditions, such as interventions~\citep{peters17, imbens15}. Querying an LLM about text interventions places it in a low-probability regime of its training data, where recent work suggests LLM reasoning struggles~\citep{udandarao24}. 

To integrate LLMs into causal inference without relying on their limited causal reasoning, new estimation methodologies are needed. Double machine learning (DML) focuses on controlling overfitting bias via sample splitting and orthogonalization~\cite{chernozhukov2018double}, and has been applied on texts and LLMs~\cite{veljanovski2024doublelingo,ma2025llm}. Another promising direction is counterfactual data generation~\citep{hamilton22, mishra2024llm, bhattacharjee2024zero, li2023prompting}. In our work, we leverage LLMs in two ways: (1) imposing interventions on text (e.g., modifying sentiment in text) without requiring causal reasoning, and (2) generating counterfactual outcomes to construct evaluation datasets.

\paragraph{Domain Adaptation}
Pre-training followed by fine-tuning on specific tasks has become a standard paradigm in language modeling~\citep{devlin2018bert, brown20, dubey2024llama}. However, fine-tuning often leads to performance degradation on unseen data due to domain shift between training and test data~\citep{ma-etal-2019-domain, ryu2022knowledge, bețianu2024dallmi}. To mitigate this, domain adaptation techniques have been developed, including sample reweighting~\citep{li2016prediction}, structural correspondence learning~\citep{blitzer-etal-2006-domain}, joint distribution matching~\citep{6751384}, and mixture of experts~\citep{guo-etal-2018-multi}. A notable approach is Domain Adversarial Neural Network (DANN)\citep{ganin2015unsupervised}, which integrates a domain classifier as an adversary to encourage domain-invariant features, improving adaptation. This method has been effective in NLP tasks such as 
stance and morality detection~\citep{allaway2021adversarial,guo2023data}. 


Domain adaptation helps address the covariate shift in causal estimation, by aligning the distributions of treated and control groups~\cite{abdullahi2021analysis}. Some classic causal methods to adjust for confounders, such as inverse propensity weighting (IPW)~\cite{hirano2001estimation} and doubly robust (DR)~\cite{funk2011doubly} methods, can also be viewed as a form of domain adaptation by aligning the distributions, and have been used to deal with incomplete data~\cite{seaman2018introduction}.
However, both IPW and DR specifically consider discrete (e.g. binary) treatments that are observed separately from the text, but have not yet been developed for the identification setting where the interventions acting directly on a piece of text.
In addition, advanced deep learning-based adaptation remains underexplored in causal estimation. In our work, we evaluate IPW, DR and DANN for predicting potential outcomes.


\section{Methods} \label{sec:methods}

\begin{figure*}[h]
    \centering
    \includegraphics[width=1\linewidth]{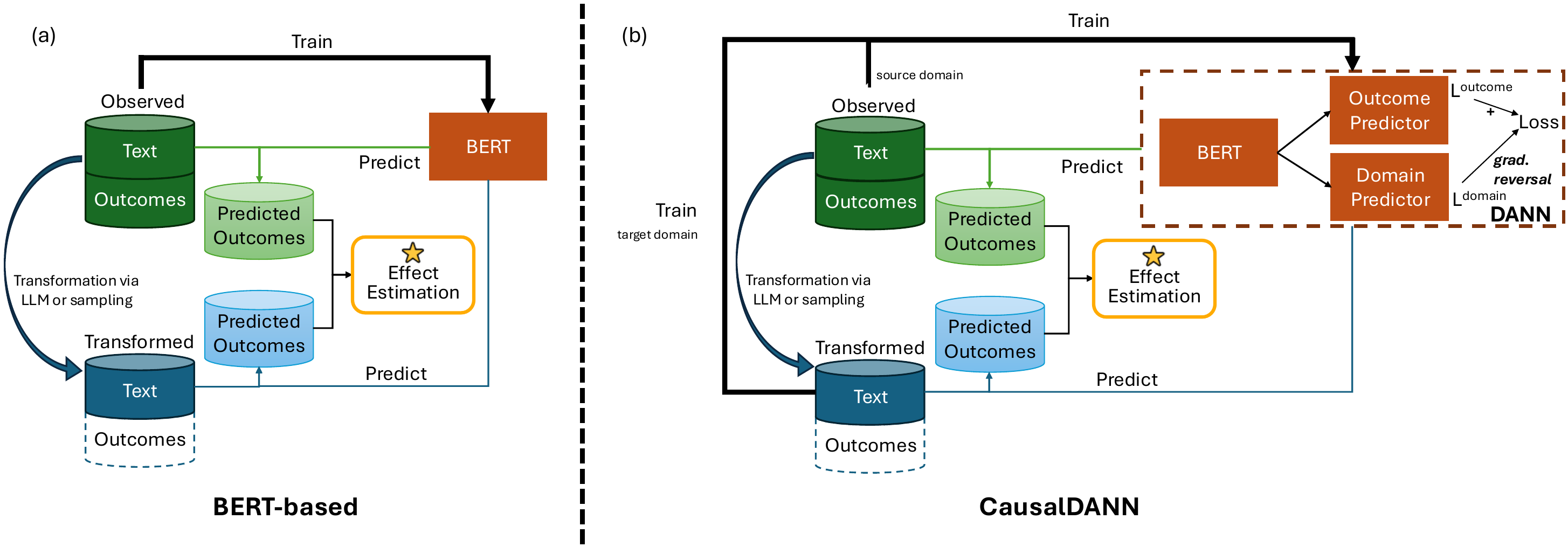}
    \caption{We first apply an LLM transformation or sampling to the observed text and outcome (non-intervened group) to generate text data for the intervened group. The outcomes for the transformed data remain unobserved. To predict the outcomes, we use (a) the BERT-based baseline predictor or (b) the proposed \textsc{CausalDANN} with domain adaptation. We then predict outcomes for both groups and compute the causal effects.}
    \label{fig:method}
\end{figure*}

We tackle the challenge of causal estimation for textual data where the treatment variable (e.g., emotion) is unobserved, and constructing the treated or control group is difficult. We propose \textbf{\textsc{CausalDANN}} (Figure~\ref{fig:method}), a framework that (1) applies a transformation on observed text to construct the intervened group, (2) predicts potential outcomes using a domain-adaptation model trained on non-intervened data, and (3) estimates causal effects from the predicted outcomes.

\subsection{Identification}
We formalize our inference problem on text with \emph{potential outcomes}~\cite{imbens15, rubin74}. For text $W$ and outcome $Y$, we construct causal estimands through an intervention on the text defined by a transformation $W\mapsto g(W)$ (e.g., rephrasing text to express more anger while preserving all other aspects). This intervention implicitly defines the treatment and results in two groups: the observed and the intervened. The causal estimand now becomes the difference in predicted outcomes between $W$ and $g(W)$. Structurally, this setup mirrors the binary treatment framework, ensuring that three key causal identification assumptions still hold:

\textbf{SUTVA} (Stable Unit Treatment Value Assumption): a unit's potential outcomes are only a function of its own text and not that of any other units.

\textbf{Overlap}: For each original text $W$, the transformed text $g(W)$ lies within the support of the \emph{representation space} covered by the observational data. We assume that text $W$ encodes both the treatment implicitly defined through the intervention, and latent covariates $C$ (e.g., grammar). When conceptualizing text in a high-dimensional representation space, the treatments and the latent covariates can be thought to correspond to some of these dimensions. A transformation $g(\cdot)$ then moves the text along specific dimensions. Overlap requires that $g(W)$ does not lie in a separate region from the observed data $W$. This enables generalization, especially through domain adaptation.

\textbf{Ignorability}: Conditional on the observed text $W$ and external covariates $X$, the treatment assignment is independent of potential outcomes. Here we distinguish between the two types of covariates---$X$ includes observed covariates external to the text (e.g., social platform where text is posted), whereas $C$ represents latent covariates of interest that are recovered from the text (e.g., grammar). See Fig.~\ref{fig:DAG}. In our estimation, we always condition on text, which contains the latent covariates $C$. This reduces the need of externally identifying and controlling for $C$. 
Both C and X can be confounding or non-confounding.

These assumptions enable the identification of potential outcomes by conditioning on the treatment text $w$, which can be observed or transformed.

{\small
\begin{align*}
    \E[Y(w)\mid X=x] = \E[Y\mid W=w, X=x] \numberthis \label{eq_assum}
\end{align*}}

We can thus compute the average treatment effect (ATE) and the conditional average treatment effect (CATE) as

{\small
\begin{align*}
    ATE &= \E[Y(g(w))] - \E[Y(w)] \numberthis \label{eq_ate}\\
    CATE &= \E[Y(g(w)) | X] - \E[Y(w) | X] \numberthis \label{eq_cate}
\end{align*}}




\subsection{Interventions on Text using an LLM} \label{sec:method_intervening}


We introduce different kinds of transformations $W\mapsto g(W)$ on observed text $W$ to generate data in the intervened group.
\begin{enumerate}
    \item 
\textbf{LLM-based transformation:} 
An LLM is prompted to rephrase text to intensify or attenuate an attribute such as anger or sentiment, while preserving all other properties. This enables targeted interventions and supports estimation of specific treatment effects (Section~\ref{sec:aita_anger}).
\item 
\textbf{Structured sampling:}
In datasets with structured formats (e.g., product reviews with ratings or Reddit posts with comments), we construct control/intervened groups by selecting alternative \emph{observed} examples---e.g., 5-star vs. 1-star reviews (Section~\ref{sec:amazon_reviews}), or top-ranked vs. randomly selected comments on the same post (Section~\ref{sec:aita_comments}).
\end{enumerate}

It may be questioned whether an intervention defined through an LLM with a certain prompt really disentangles and modifies one aspect of language while preserving the others. LLMs' capabilities in counterfactual generation are well-known~\cite{brown20,dubey2024llama,mishra2024llm,li2023prompting}; however, they are not free of all social bias. We conduct manual inspections and analyze potential biases in our data and did not observe any significant bias (see Appendix~\ref{sec:biases}). Future applications of our framework should validate the fidelity of specific LLM-defined transformations using techniques such as improved prompting~\cite{li2023prompting}, human-in-the-loop annotations, and improved alignment. Our work studies how to estimate the effect of a given intervention.

\subsection{Counterfactual Generation with LLMs}Even after applying transformation techniques to construct the intervened group, outcomes remain unobserved, and ground truth causal effects are unavailable. Thus, generating synthetic data is crucial for evaluation. LLMs, with their extensive training and reinforcement learning from human feedback (RLHF), can simulate outcomes for unobserved data, such as modeling social norms~\citep{brown20,dubey2024llama}. 
In Sections~\ref{sec:aita_comments} and~\ref{sec:aita_anger}, we use LLMs to simulate social judgments on posts from Reddit’s /r/AmITheAsshole. Although using LLMs for counterfactual generation in evaluation data may introduce biases, we did not find any in our case studies (see analyses and mitigation strategies in Appendix~\ref{sec:biases}).

\subsection{Outcome Prediction}
Since the outcomes of the generated (intervened) data are never observed, we need a prediction model to estimate the outcomes for all the data in order to estimate the causal effects. \textsc{CausalDANN} improves robustness to the domain shift between observational and intervened text by using domain adversarial training to learn domain-invariant features, enabling more accurate prediction on transformed texts.

\subsubsection{BERT} \label{sec:methods_BERT}
BERT based models~\citep{devlin2018bert} are among the best approaches for adjusting for textual confounding~\cite{feder2021causalm,veljanovski2024doublelingo,keith2020text}. Thus, we use it as our \emph{baseline} outcome predictor. 
We train a vanilla BERT model using only non-intervened text, where outcomes are observed and serve as labels for supervised learning. To mitigate the bias from BERT predictions being propagated into effect estimation, we predict the outcomes for both control and treated with the same BERT mode, ensuring a fairer comparison when estimating effects as a relative difference between the two groups.


We build the outcome predictor by appending a linear classification layer to the BERT embedding model (see Appendix~\ref{sec:model_training}). To obtain the predicted outcome for all data, we randomly split the data into five folds. Each time we train the outcome predictor with four-fifths of the data, which is further split into training and validation sets by 80\%-20\% ratio. Finally, we predict on the remaining one-fifth data to obtain their predicted outcomes. Given the $i$-th text $W_i$ and its observed covariate $X_i$, the outcome can be predicted as ${Y_i} = E[Y|W,X] \approx \mu_{BERT}(W_i,X_i)$ from the BERT-based classification model. The ATE can be estimated as

{\small
\begin{align*}
\small
ATE 
&= \E[Y(g(W))] - \E[Y(W)] \\
&\approx \frac{1}{N} \sum_{i=1}^{N} \mu_{\text{BERT}}(g(W_i), X_i) - \mu_{\text{BERT}}(W_i, X_i) \numberthis
\end{align*}}

where $N$ is the total number of data points and $g(\cdot)$ is the text transformation function.
The CATE can be computed by using equations~\ref{eq_assum} and \ref{eq_cate}

{\small
\begin{align*}
CATE &= \E[Y(g(W)) | X] - \E[Y(W) | X]\\
&= \E[ \E[Y | W=g(W), X] | X] - \E[Y(W) | X] \numberthis
\end{align*}}

We can use BERT to estimate the inner expectation $\E[Y | W=g(W), X] \approx \mu_{BERT}(g(W),X)$ in the first term, and the outer expectation marginalizes out the W. The second term can be similarly computed with $\mu_{BERT}(W,X)$.

In CATE, the confounding external covariates $X$ needs to be conditioned on, but the latent covariates $C$ is not explicitly necessary for causal identification, since we always condition on text. In conventional settings (e.g. $T$ is anger), adjusting for latent language properties $C$ (e.g. grammar) is difficult, but our framework reduces this need. Nevertheless, non-confounding features can also be used to analyze heterogeneous treatment effects. For example, while topic ($C$) is controlled in LLM transformation and not a confounder, we can still examine effects across different topics.

{\small
\begin{align*}
\small
CATE &\approx\frac{1}{N_{c,x}}\sum_{i:C_i=c,X_i=x}^{N_{c,x}}\mu_{BERT}(g(W_i),X_i)\\
&- \mu_{BERT}(W_i,X_i)\ \ \forall c \in \mathcal{C}, x \in \mathcal{X} \numberthis
\end{align*}}

\subsubsection{\textsc{CausalDANN}} 
Fine-tuned BERT often suffers performance drops on unseen data~\cite{ma-etal-2019-domain,ryu2022knowledge}. Since our outcome predictor must perform well on both non-intervened data with observed outcomes and transformed data  with unobserved outcomes (section~\ref{sec:method_intervening}), we adopt Domain Adversarial Neural Network (DANN) instead of vanilla BERT. \citet{guo2023data} demonstrated DANN’s strong performance in domain adaptation for textual data. 

DANN mitigates data shift by mapping text embeddings from both labeled source and unseen target domains onto a shared space. \textsc{CausalDANN} consists of three modules (Figure~\ref{fig:method}b): (1) a BERT encoder for textual representation, (2) an outcome predictor---a linear classifier same as in the vanilla BERT outcome predictor, and (3) a domain predictor---another linear classifier but with cross-entropy loss trained adversarially to be maximized, ensuring domain-invariant embeddings.
This is achieved by connecting the domain classifier to the other parts of the model with a gradient reversal layer. The loss term is:
\begin{equation} \label{eq:DANN_loss}
    L = L^{outcome} - \lambda^{D} \cdot L^{domain}
\end{equation}
where $\lambda^{D}$ is a loss-balancing hyperparameter (see training details in Appendix~\ref{sec:model_training}).

In our setup, the source domain is non-intervened (observed) data, while the target domain is intervened (unobserved) data. We train the model in a semi-supervised way, providing both the labeled non-intervened training data and the unlabeled intervened data in each batch, balanced in size. Both pass through the BERT encoder to learn textual representations. Next, the non-intervened data go through both the outcome and domain predictors, while intervened data, lacking outcome labels, only pass through the domain classifier. This adversarial setup pushes the encoder to learn domain-invariant embeddings, aligning intervened data closer to the non-intervened labeled data (see Appendix~\ref{sec:model_training}). After obtaining predicted outcome by \textsc{CausalDANN}, the ATE and CATE can be caculated in the same way as in Section~\ref{sec:methods_BERT}.


\section{Experiments} \label{sec:experiments}

We use real-world data from Amazon product reviews~\cite{ni-etal-2019-justifying} and Reddit r/AmITheAsshole discussions of social dilemmas~\footnote{\url{https://github.com/iterative/aita_dataset}} for model evaluation in three case studies.

\subsection{Baselines and Setups}
We compare three baselines (1) BERT, (2) inverse propensity weighting (IPW) and (3) doubly robust estimator (DR) against the proposed (3) \textsc{CausalDANN}. We also evaluate (4) \textsc{CausalDANN}+DR to assess whether DR further improves domain adaptation. See details of baseline implementations in Appendix~\ref{sec:model_training}. Additionally, we compare with (5) TextCause~\cite{pryzant21}, a causal estimation method that, like ours, adjusts for confounding using BERT embeddings. However, TextCause assumes observations exist in both control and treatment groups, 
and fails when the treatment variable is unobserved or when one group lacks data. Nevertheless, The TextCause performance tells us how good the effect estimation can get to even with more training data from both control and treated groups.

We evaluate performance using two metrics: (1) $\Delta ATE$, the absolute difference between predicted and ground truth ATE, and (2) mean squared error (MSE) of the CATE vector across all covariate dimensions. For both, lower values indicate better performance. Each experiment is repeated five times to report the average and standard deviation.

\subsection{Amazon Reviews} \label{sec:amazon_reviews}

\subsubsection{Data and Experiment Set-up}
We first evaluate our method on a semi-synthetic dataset based on Amazon reviews~\cite{ni-etal-2019-justifying}, a benchmark also used in~\cite{pryzant21}. This dataset consists of 5.6K reviews on products in the categories of mp3, CD, or Vinyl. Reviews for products worth more than \$100 or fewer than 5 words are excluded.

Following~\cite{pryzant21}, we estimate the effect of positive sentiment in reviews on sales, hypothesizing that positive reviews increase click and purchase likelihood, though product type may confound this effect. See causal graph in Figure~\ref{fig:DAG_amazon}. Since sentiment is embedded in text and not explicitly observed, defining the treatment variable in a conventional causal setting is challenging.

Based on our framework, we define a transformation function $g(\cdot)$ that intervenes on the positive sentiment of the reviews without manipulating other attributes like grammar. In this experiment, we do not use an LLM, but instead we sample the positive and non-positive reviews from the observed data itself, labeling reviews as positive if they have 5 stars and non-positive if they have 1 or 2 stars. The intervention is $T=\mathbbm{1}_{positive}$. This setting (1) evaluates our framework without potential biases from LLM-generated text, and (2) mirrors a traditional causal estimation setup, enabling a fair comparison with TextCause.

The observed covariate $X$ is a binary indicator for whether the associated review is a CD or not. The outcome $Y$, a binary variable for whether a product received a click or not, is simulated based on the textual intervention $T=\mathbbm{1}_{positive}$ and the covariate $X$ as following:
\begin{equation} \label{eq_amazon_gen}
    Y \sim \text{\small Bernoulli}(\sigma(\beta_{X}(\pi - \beta_0) + \beta_TT + \varepsilon))
\end{equation} 
where $\sigma(\cdot)$ is the sigmoid function, $\pi=P(T|X)$ is the propensity, $\beta_X$ controlling confound strength is set to 4.0, $\beta_T$ controlling treatment strength is set to 0.8, $\beta_0$ controlling propensity is set to 0.8. All these hyperparameters are set as the same as a harder-to-estimate scenario in~\cite{pryzant21}.

We then follow our framework and train the outcome and propensity predictors. We assume that we do not observe the outcomes in the intervened (positive) group. Therefore, we only use the non-intervened reviews and their labels $Y(T=0)$ to train the outcome predictors. For the propensity predictor in the IPW method, we use $T$ as the training labels and both intervened ($T=1$) and non-intervened ($T=0$) groups.

\paragraph{Identification} The outcome generation process (Equation~\ref{eq_amazon_gen}) ensures the SUTVA, overlap and ignorability assumptions.

\subsubsection{Results}

Table~\ref{tab:amazon_results} shows that \textsc{CausalDANN} gives us the ATE and CATE estimation closest to the ground truth, outperforming other baselines including BERT, IPW, DR and also \textsc{CausalDANN}+DR. The fact that \textsc{CausalDANN} achieves lower error than the vanilla BERT baseline tells us the effectiveness of domain adaptation by DANN. In addition, the results show that IPW has significantly worse performance, suggesting issues with propensity score estimation. We notice that the propensity predictor easily classifies reviews as positive or negative, yielding probability outputs near 0 or 1, even after temperature scaling. This extreme weighting leads to the numerical ill behavior in IPW. DR also utilize the propensity score. For the same reason, DR also has worse performance than \textsc{CausalDANN}, although its bias is much less significant than IPW due to DR's doubly de-biasing approach. Last, TextCause sets an upper bound on performance by leveraging more observations in the intervened group. It is no surprise that TextCause performs better. However, \textsc{CausalDANN} achieves a $\Delta ATE$ close to TextCause and matches its MSE for CATE, demonstrating that its estimations are less biased.

\begin{table}[h]
    \centering
    \resizebox{0.9\linewidth}{!}{
    \renewcommand{\arraystretch}{1.2}
    \begin{tabular}{l|c|c}
    \Xhline{1pt}
         & \textbf{$\Delta ATE$  ($\downarrow$)} & \textbf{MSE of CATE  ($\downarrow$)} \\
         \hline
        TextCause & $\mathbf{0.05 \pm 0.02}$ & $\mathbf{0.01 \pm 0.01}$ \\
        \hline
        BERT & $0.12 \pm 0.05$ & $0.02 \pm 0.01$ \\
        IPW & $27.29 \pm 6.22$ & $1135.02 \pm 416.02$ \\
        DR & $0.58 \pm 0.42$ & $1.26 \pm 1.27$ \\
        \textsc{CausalDANN} & $\mathbf{0.09 \pm 0.04}$ & $\mathbf{0.01 \pm 0.01}$ \\
        \textsc{CausalDANN}+DR & $1.37 \pm 0.40$ & $2.99 \pm 1.30$ \\
    \Xhline{1pt}
    \end{tabular}}
    \caption{Causal estimation errors in Amazon Reviews.}
    \label{tab:amazon_results}
\end{table}

\subsection{Reddit AITA Comments} \label{sec:aita_comments}

\subsubsection{Data and Experiment Set-up}
Next we evaluate our methods on the Reddit r/AmITheAsshole (AITA) data, where users share social dilemmas and receive judgements on who is at fault. This data has been used as a benchmark for causal estimation~\cite{marmarelis24_ensembled}.

In AITA, the verdict is shaped by upvotes on comments, with the most upvoted comment remaining at the top, increasing its visibility. A key research question is whether exposure to the top comment affects individual judgment, due to influences of mainstream opinions. Conducting such an experiment in real life is challenging, but our framework enables an approximation to this causal question utilizing LLM-generated synthetic evaluation data. See causal graph in Figure~\ref{fig:DAG_aita_comments}.

In this experiment, we do not use LLM transformation as an intervention, but sample both intervened and control units from real-world data. We select the top-ranked comment on each Reddit post as the intervened unit and randomly sample another comment (excluding the top-ranked one) on the same post as the control unit. The intervention is thus the binary variable $T = \mathbbm{1}_{top\ comment}$.

To create a semi-synthetic evaluation dataset, we perform LLM counterfactual generation by prompting GPT-4~\footnote{\url{https://openai.com/index/gpt-4/}} to act as a moral judge on these real-world posts and generate verdicts~\cite{marmarelis24_ensembled}. This simplifies the problem from estimating effects on real-life human behavior to analyzing impact in an LLM-simulated scenario.  See Appendix~\ref{sec:appendix_prompting} for prompting procedure. Table~\ref{tab:aita_comment_generations} shows example generations, which exhibit good quality.
The outcome Y is simulated by prompting an LLM to provide moral verdicts for (a) a story with its top-ranked comment and (b) the same story with a randomly selected comment. We formulate outcome $Y$ to be binary, with $Y=1$ when GPT assigns fault and $Y=0$ otherwise (we categorize YTA (you are the asshole) and ESH (Everyone sucks here) into $Y=1$, and NTH (not the asshole) and NAH (no asshole here) into $Y=0$). For each post, we generate both outcomes $Y(T=1)$ and $Y(T=0)$ to compute synthetic ground truth treatment effects. We examine potential LLM generation biases in Appendix~\ref{sec:biases}.

Following the approach in Section~\ref{sec:methods}, we train the outcome predictor and the propensity predictor. Both predictors use concatenated post and comment text, adjusting for latent confounders in texts.
The outcome predictor is trained using only non-intervened data with observed ground truth $Y(T=0)$ labels, while the propensity predictor uses $T = \mathbbm{1}_{top\ comment}$ labels from both groups. 
Additionally, we compute CATE conditioned on post topics, which we extract using BERTopic~\cite{grootendorst22}. See Appendix~\ref{sec:appendix_bertopic}.

\paragraph{Identification}
We use LLM counterfactual generation to simulate pseudo ground truth outcomes for model evaluation purpose. In other words, we estimate the average difference in the GPT judge’s (instead of real humans’) outputs given an observational top comment versus a randomly picked comment in its place. This outcome simulation process ensures ignorability, as only the comment changes while all other aspects (e.g., the post) remain the same. We assume overlap is also satisfied. Our covariates are latent in the post and comment text. Each pair of intervened and control units have the same post. We assume comments are also similar in style, because both top and random comments originate from the same discussion thread. Thus, as explained in Section~\ref{sec:methods}, observational and intervened texts should occupy the same high-dimensional space except along the "treatment dimension". Simulating each verdict in an independent process partially ensures SUTVA, and we acknowledge that the simulated outcomes could still be correlated across different units because they are generated by the same LLM model, which is a limitation of how this testbed is constructed. We hope to add more alternative evaluations in future works. 

\subsubsection{Results}

\begin{table}[h]
    \centering
    \resizebox{0.9\linewidth}{!}{
    \renewcommand{\arraystretch}{1.2}
    \begin{tabular}{l|c|c}
    \Xhline{1pt}
         & \textbf{$\Delta ATE$ ($\downarrow$)} & \textbf{MSE of CATE ($\downarrow$)}\\
        \hline
        TextCause & $0.04 \pm 0.01$ & $0.01 \pm 0.00$ \\
        \hline
        BERT & $\mathbf{0.01 \pm 0.01}$ & $0.01 \pm 0.01$ \\
        IPW & $3.43 \pm 1.02$ & $14.08 \pm 9.03$ \\
        DR & $0.07 \pm 0.03$ & $0.04 \pm 0.01$ \\
        \textsc{CausalDANN} & $\mathbf{0.01 \pm 0.01}$ & $\mathbf{0.00 \pm 0.00}$ \\
        \textsc{CausalDANN}+DR & $0.21 \pm 0.06$ & $0.08 \pm 0.03$ \\
    \Xhline{1pt}
    \end{tabular}}
    \caption{Causal estimation errors in AITA comments.}
    \label{tab:aita_comment_results}
\end{table}

Table~\ref{tab:aita_comment_results} shows that \textsc{CausalDANN} performs exceptionally well on this dataset. It has similar performance to BERT because top and random comments are often similar in perspective and writing style, leading to minimal data shift between non-intervened and intervened data. As a result, BERT also performs well here. The similarity between the top and random comments also leads to inaccurate propensity score estimation. Therefore, IPW and DR-based methods perform worse due to low accuracy in the propensity prediction (F1-score is $0.52 \pm 0.01$). Finally, although TextCause cannot handle unobserved outcomes in the intervened group, we run it by providing LLM-generated counterfactuals. We find \textsc{CausalDANN} slightly outperforms TextCause using more information.

\subsection{Anger in AITA Posts} \label{sec:aita_anger}

\subsubsection{Data and Experiment Set-up}
In the same AITA dataset, we investigate whether the anger level in posts affects people's verdicts. Posts with higher anger or aggression may lead to a higher likelihood of being deemed the author’s fault. See Figure~\ref{fig:DAG_aita_anger} for causal graph. Anger is a latent attribute in texts. Under a conventional setup, we would need to use a proxy treatment variable which could be susceptible to more bias~\cite{pryzant21}. In our framework, we define an intervention $T = \mathbbm{1}_{anger}$, where we use Claude 3.5 Sonnet~\footnote{We use Claude 3.5 Sonnet (\url{https://www.anthropic.com/news/claude-3-5-sonnet}), as GPT is more conservative and does not significantly alter the anger level.} to transforms each post to increase its anger level while preserving style, semantics, and perspective. To mitigate bias from LLM transformations, we also rephrase the original post keeping the same anger level. Manual inspection confirms the quality of the transformations (details in Appendix~\ref{sec:appendix_prompting}). 

Similar to section~\ref{sec:aita_comments}, we create a semi-synthetic evaluation data by counterfactual generation of outcomes. 
$Y$ is generated by prompting an LLM for verdicts on (a) an LLM-anger-transformed AITA story and (b) an LLM-rephrased version of the original story. We are estimating the effect of the intervention formulated as this LLM-anger-transformation, conditioned on text containing latent text-related covariates and the implicit treatment. More analysis on potential bias from LLM-generations, data shift and overlap from LLM-transformation can be found in Appendix.

Similar to previous experiments, we only use the non-intervened (rephrased) posts and their corresponding Claude-generated verdicts to train the outcome predictor. To train the propensity predictor, we use all data and $T = \mathbbm{1}_{anger}$ as the label. The input to both predictors are the post texts. In addition, we also have the covariates $\mathbf{C}$ to be the binary vector of the top 30 frequent topics obtained from BERTopic for computing CATE. The training procedure is similar to previous experiments (Appendix~\ref{sec:model_training}).

\paragraph{Identification} Similar to the previous experiment, all confounding latent covariates are controlled, and ignorability and overlap assumptions are satisfied. Running each verdict simulation independently partially ensures SUTVA, though simulated outcomes may still correlate since they come from the same LLM. We plan to explore additional evaluation methods in future work.

\subsubsection{Results}

From Table~\ref{tab:aita_anger_results}, we see that \textsc{CausalDANN} outperforms other methods and matching the performance of TextCause (using more training data from the intervened group). Methods using DANN architecture are better than methods based on vanilla BERT, again showing the effectiveness of domain adaptation in predicting potential outcomes. Applying IPW again signifincantly lowers the performance because of the numerical ill behavior when the propensity scores were close to either 0 or 1. \textsc{CausalDANN} also outperforms DR again in this data. All these results indicate that propensity estimation, IPW and DR can be less robust. 

\begin{table}[h]
    \centering
    \resizebox{0.9\linewidth}{!}{
    \renewcommand{\arraystretch}{1.2}
    \begin{tabular}{l|c|c}
    \Xhline{1pt}
         & \textbf{$\Delta ATE$ ($\downarrow$)} & \textbf{MSE of CATE ($\downarrow$)} \\
        \hline
        TextCause & $0.05 \pm 0.01$ & $0.02 \pm 0.01$ \\
        \hline
        BERT & $0.09 \pm 0.07$ & $0.03 \pm 0.02$ \\
        IPW & $154.61 \pm 16.24$ & $25744.42 \pm 5467.05$ \\
        DR & $0.10 \pm 0.05$ & $0.31 \pm 0.36$ \\
        \textsc{CausalDANN} & $\mathbf{0.05 \pm 0.03}$ & $\mathbf{0.01 \pm 0.01}$ \\
        \textsc{CausalDANN}+DR & $0.07 \pm 0.04$ & $0.15 \pm 0.18$ \\
    \Xhline{1pt}
    \end{tabular}}
    \caption{Causal estimation errors in AITA-anger data.}
    \label{tab:aita_anger_results}
\end{table}


\section{Conclusions}

In this work, we target the problem of causal estimation for textual data especially when the treatment variable is not directly observed. We propose \textbf{\textsc{CausalDANN}}, a framework for estimating the causal effect of tailored interventions on text. Using domain-adaptive text classifiers, our approach produces robust effect estimates while handling domain shifts. We demonstrate \textsc{CausalDANN}'s strong performance across three experiments.

\paragraph{Limitations} 
First, LLM-generated data is prone to biases and limited diversity (Appendix~\ref{sec:biases}). We mitigate these issues through manual inspection of the generations, and by applying LLM rephrasing to the non-intervened group alongside the transformation of intervened group. This ensures a fair comparison when estimating effects as a relative difference between the two groups. If the transformation intervention does contain biases from LLMs, then this bias would propagate to effect estimation. Our evaluation on three different datasets shows that our method achieves lower biases in causal estimation than the baselines. Bias mitigation in counterfactual generation is an active research area, including strategies like enhanced prompting and human-in-the-loop annotation~\cite{li2023prompting,bhattacharjee2024zero}. However, this topic is beyond the scope of our paper.

Second, our experiments evaluate the proposed framework using LLM-generated semi-synthetic data, meaning the estimated effects reflect simulated LLM textual transformations not real human social behaviors. However, applying this approach to causal estimation on real data requires the assumption that LLMs can reliably infer unobserved data points through text transformation based on observed human behavior---an assumption that may not always hold. For instance, although we control for linguistic properties such as semantics and grammar through prompting, modifying anger levels via LLM transformation may also unintentionally alter toxicity. Future applications of our framework should also carefully validate this assumption. We also plan to explore fine-tuning LLMs with observed data, enabling them to learn specific personas and more closely align their behavior with human social systems. Alternative testing methods can also be designed in the future. For example, in AITA-comments experiment, we can change the prompt so that GPT is given the same comment twice but told once that it was a top-ranked comment and once that it was a low-ranked comment.

Third, the consistency assumption requires that there are no two ‘flavors’ or versions of treatment~\cite{rehkopf2016consistency}. However, we define the treatment T as a transformation operation $T = W \rightarrow g(W)$, which is different from the conventional binary treatment. We therefore interpret potential outcomes as defined relative to the sampling/transformation procedure. The experiments with LLM-constructed semi-synthetic data can violate the strict consistency assumption in conventional setup. For example, in the AITA-anger experiment, there may be different ``flavors'' of the anger treatment as there are sources of randomness in the LLM sampling procedure. We propose to conduct more rigorous estimation in future works, such as studying multiple random generations for one input text. In LLM-based rewrites, we also mitigate the randomness in generation via controlled prompting and rephrasing baselines.

Last, there can be potential unobserved confounders that are not accounted for in our experiments. While we adjust for confounding by conditioning on textual representations and estimated propensity scores, there may still be latent factors influencing both the treatment and outcome that our models fail to capture. For example, in the AITA dataset, implicit biases in user interactions in the Reddit threads could affect verdicts in ways not reflected in the observed text. Addressing unobserved confounding remains a fundamental challenge in causal inference from observational data, and future work could explore the proper sensitivity analysis to bound the uncertainty.

\paragraph{Ethics Statement}
In this research, we utilized Large Language Models (LLMs) to generate data for analysis. We acknowledge the ethical considerations associated with the use of LLMs, particularly concerning potential biases, data validity, and the broader implications of AI-generated content. The raw Reddit AITA and Amazon reviews data we use contain information like names or uniquely identifies individuals, and contain offensive content. All the user names are anonymized to ID strings. Additionally, we only present the aggregated average effects, without revealing any individual information. To maintain transparency, we have detailed the methodology used for data generation and the steps taken to address ethical concerns in this paper. We encourage readers to consider these factors when interpreting our findings. Furthermore, our causal estimation framework, while effective in controlled semi-synthetic settings, may not generalize perfectly to real-world scenarios. In real data applications, unobserved confounders and biased treatment effect estimations could lead to misleading conclusions. Users of our method should be cautious when applying it to real-world interventions, particularly in high-stakes domains such as policy-making, healthcare, or legal decision-making, ensuring appropriate validation and sensitivity analyses.

Finally, this manuscript was drafted by the authors with the assistance of ChatGPT to refine the language and improve readability. All content was reviewed and verified for accuracy.



\bibliography{refs}

\appendix

\section{Causal Graphs and Identification in Three Experiments} \label{sec:appendix_identification}

\paragraph{Amazon Reviews} We estimate
the effect of positive sentiment in reviews on sales,
hypothesizing that positive reviews increase click
and purchase likelihood, though product type may
confound this effect. In this causal question, we assume the treatment (positive or negative rating) $T=\mathbbm{1}_{positive}$ affects the review text. T affects the outcome of whether the product gets a click. We also assume the confounding covariate of product type $X$ affects both the treatment and the outcome. There might be some latent non-confounding covariates such as grammar that affects the review text  and the outcome $Y$ too. However all these latent $C$ are already taken into account in the BERT embedding during modeling process. The outcome generation procedure based on equation~\ref{eq_amazon_gen} makes sure that all three identification assumptions, SUTVA, overlap and ignorability are satisfied.

\begin{figure}[h]
    \centering
    \includegraphics[width=1\linewidth]{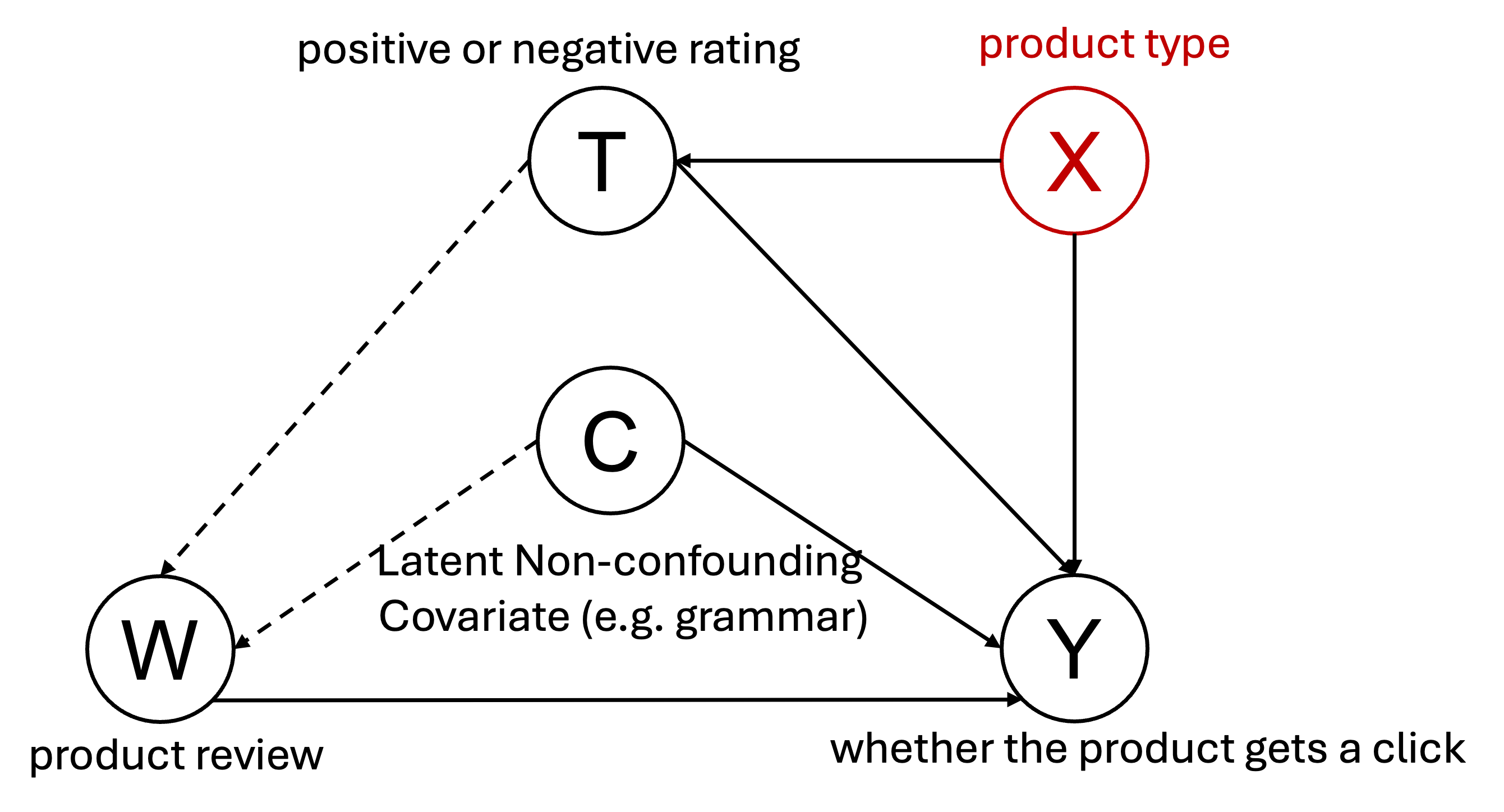}
    \caption{The assumed causal graph for the Amazon Reviews Experiment.}
    \label{fig:DAG_amazon}
\end{figure}

\paragraph{Reddit AITA Comments}
We estimate whether exposure to the top/random comment affect GPT's moral judgment, as we are using simulated GPT verdicts as the synthetic ground truth counterfactuals for model evaluation in this experiment. The treatment $T$ (top or random comment) affects the outcome, GPT's verdict. We do not consider the treatment affects the text $W$, which is the post itself. We see post topic as a latent confounding covariate. Both topic and other non-confounding latent covariates are adjusted in the BERT embeddings already. For research interest, we still use BERTopic to categorize the topic of each post and measure the heterogeneous effects.

We use LLM counterfactual generation to simulate pseudo ground truth outcomes for model evaluation purpose. In other words, we estimate the average difference in the GPT judge’s (instead of real humans’) outputs given an observational top comment versus a randomly picked comment in its place. This outcome simulation process ensures ignorability, as only the comment changes while all other aspects (e.g., the post) remain the same. We assume overlap is also satisfied. Our covariates are latent in the post and comment text. Each pair of intervened and control units have the same post. We assume comments are also similar in style, because both top and random comments originate from the same discussion thread. Thus, as explained in Section~\ref{sec:methods}, observational and intervened texts should occupy the same high-dimensional space except along the "treatment dimension". Simulating each verdict in an independent process partially ensures SUTVA, and we acknowledge that the simulated outcomes could still be correlated across different units because they are generated by the same LLM model, which is a limitation of how this testbed is constructed. We hope to add more alternative evaluations in future works. 

An additional note on SUTVA is that in general our framework (Section~\ref{sec:methods}) does not assume identifiability from LLM generation itself. Instead, we articulate conditions (ignorability, overlap, SUTVA) under which causal estimation is valid when treating interventions as hypothetical transformations $g(W)$. Identifiability in our framework derives from standard causal assumptions applied to interventions on text, not from assuming that an LLM can produce the true counterfactual. The use of LLMs in experiments is only to create semi-synthetic evaluation data, not to grant identifiability by fiat.

\begin{figure}[h]
    \centering
    \includegraphics[width=0.8\linewidth]{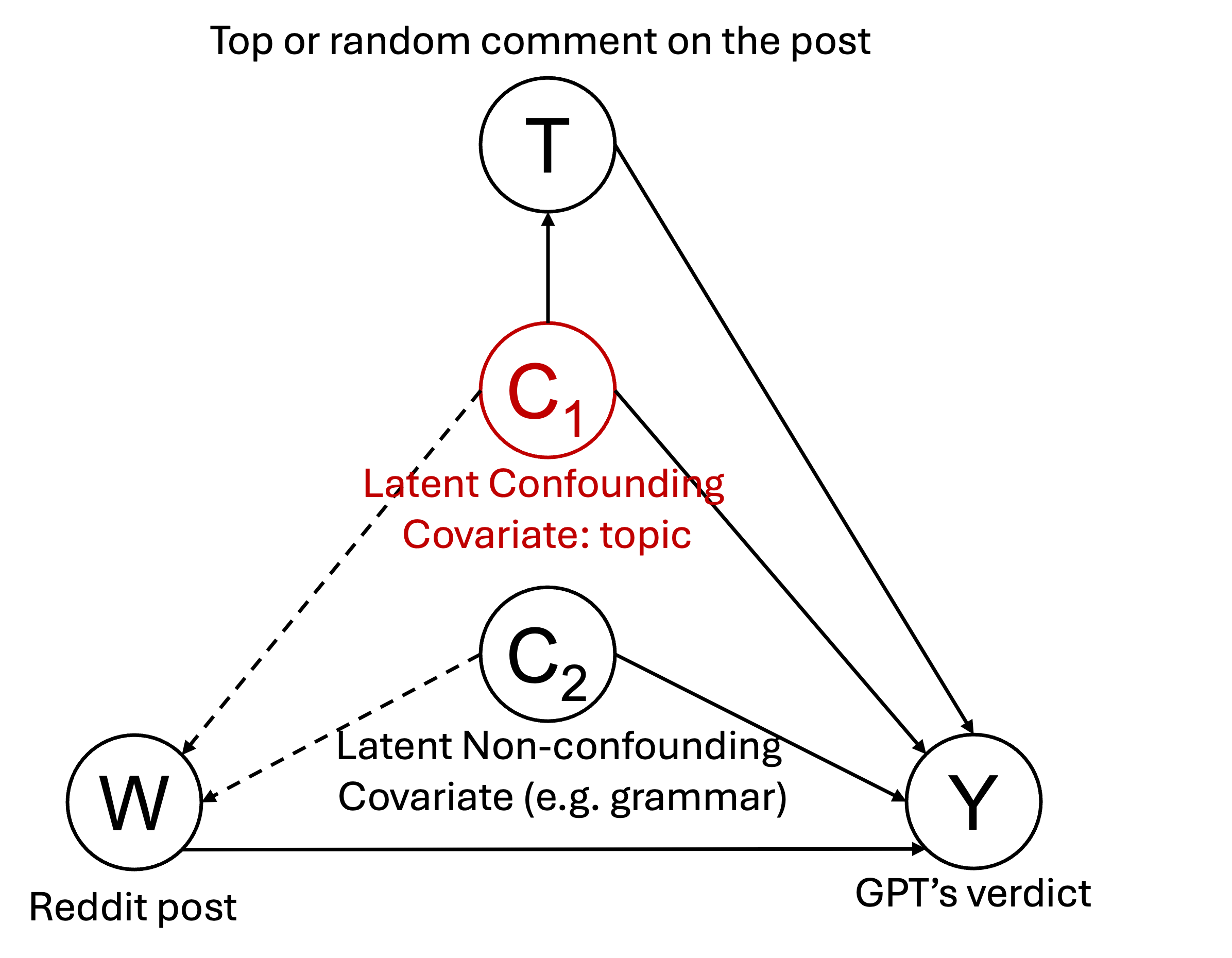}
    \caption{The assumed causal graph for the Reddit AITA Comment Experiment.}
    \label{fig:DAG_aita_comments}
\end{figure}

\paragraph{Reddit AITA Anger}
We estimate how transforming a Reddit post to an angrier version by LLM can affect GPT's moral verdict. The treatment, the LLM anger-transformation operation, affects the text $W$, and T also affects the outcome Y. Similar to the AITA Comments experiment, both topic and other non-confounding latent covariates are adjusted in the BERT embeddings already. For research interest, we still use BERTopic to categorize topic of each post and measure the heterogeneous effects. All discussions about SUTVA, overlap and ignorability assumptions are the same as in Reddit AITA Comments experiment.

\begin{figure}[h]
    \centering
    \includegraphics[width=0.8\linewidth]{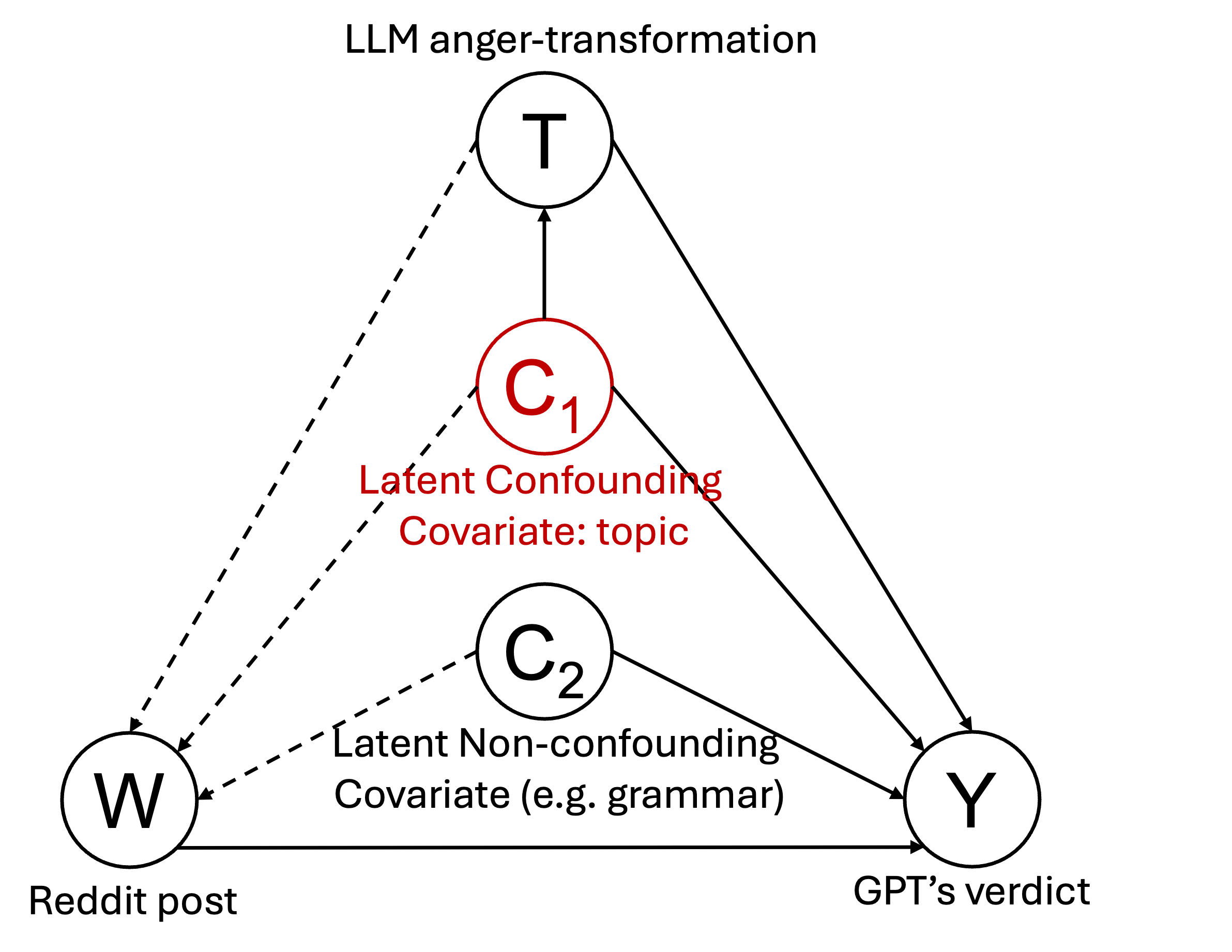}
    \caption{The assumed causal graph for the Reddit AITA Anger Experiment.}
    \label{fig:DAG_aita_anger}
\end{figure}

\section{Mitigation of LLM-related Biases} \label{sec:biases}
LLM-generated text can be biased due to imperfections in training data, affecting both (1) counterfactual generation for synthetic test data and (2) text interventions via LLM transformations.

\begin{figure}[h]
    \centering
    \includegraphics[width=1.0\linewidth]{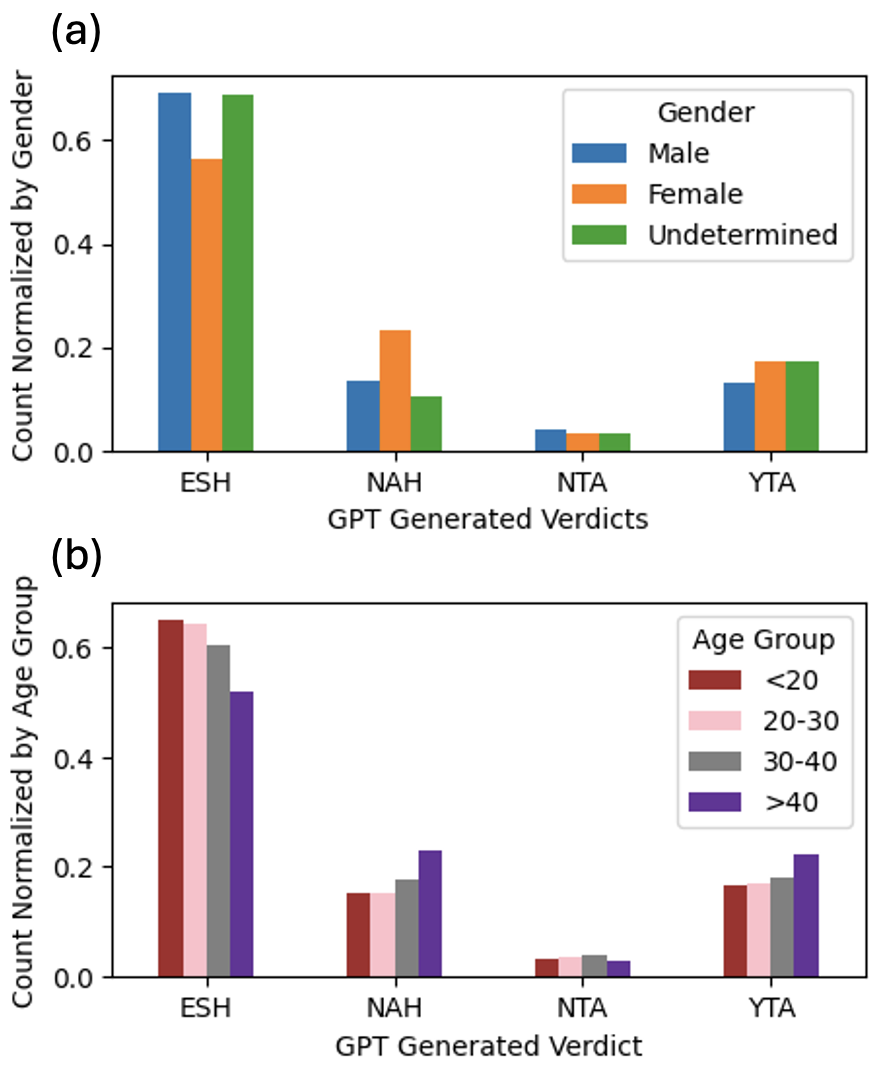}
    \caption{GPT generated AITA verdicts in different (a) age and (b) gender groups.  We use regex to capture these.}
    \label{fig:genderage_verdicts}
\end{figure}

For the first point, we assume LLMs have learned social norms and common sense from extensive online data. To assess potential biases, we analyze the distribution of LLM-generated AITA verdicts across self-reported gender and age groups, detected via regex from real posts (e.g., "I (30F)..."). We observe differences in verdict distributions by age and gender (Figure~\ref{fig:genderage_verdicts}), such as younger authors being labeled "ESH" more often and older authors receiving "NAH" more frequently. However, this may be a reflection of real-world human behaviors where people with different ages tend to talk about different types of stories, rather than reflecting LLM bias. To further test for bias, we rerun the same prompt on identical stories while randomly altering age or flipping gender. The verdict remains unchanged 91\% of the time with a random age and 93\% with a flipped gender, suggesting minimal bias in LLM-generated verdicts. In addition, we had two independent and well-trained expert annotators manually check 50 randomly selected generations to ensure the quality. See Appendix~\ref{sec:appendix_prompting} for prompting procedure and Table~\ref{tab:aita_angry_generations} for example GPT generations. Last, using GPT-generated verdicts for both treated and control data also mitigates the biases propagating into our causal estimation.

Disclaimer: we binarize gender labels here for simplicity, but we acknowledge that gender is not binary, and this simplification does not fully capture the diversity of gender identities.

With respect to the biases in interventions on text using LLM text transformation, we also manually review 50 randomly selected generations and do not observe bias present. See Appendix~\ref{sec:appendix_prompting} for prompting procedure and Table~\ref{tab:aita_comment_generations} for example GPT generations, which exhibit good quality. To further mitigate this bias, other than performing the text transformation defined as the intervention (e.g., increasing anger), we also rephrase the original textual data using an LLM. In this way, both non-intervened and intervened groups are LLM-generated. The estimated effect is a relative difference between the two groups. With text and outcomes in both groups being LLM-generated, the biases should cancel out to a large extent.

\section{Data shift and Overlap in AITA}
We provide additional results analyzing overlap and data shift in intervened and non-intervened groups. We assess the anger levels in Reddit AITA stories for both control and treated data using SpanEmo~\cite{alhuzali2021spanemo}, a BERT-based emotion classifier trained on the SemEval social media dataset, which has been validated in prior work~\cite{burghardt2024socio}. The figure below shows that original Reddit data (blue) exhibits greater variance in anger, while LLM-transformed data (yellow) predominantly contains texts with high anger probability—expected since the LLM was prompted to intensify anger. The observed data shift is consistent with the observation that CausalDANN with domain adaptation has better performance. In addition, there is still an overlap between the two distributions, allowing the model to adapt from the original to the unseen domain.

\begin{figure}
    \centering
    \includegraphics[width=1\linewidth]{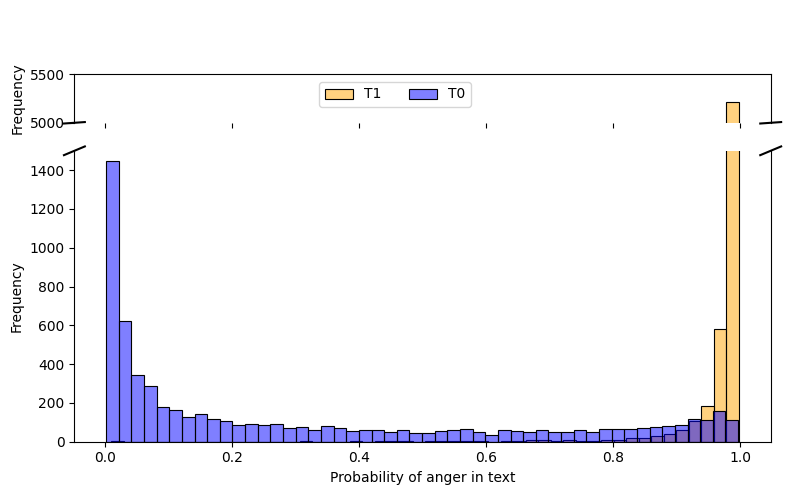}
    \caption{Distribution of Anger in control and intervened data. T0 is observed texts and T1 is LLM-transformed texts with respect to anger.}
    \label{fig:anger_dist}
\end{figure}

\section{Model Training and Hyperparameters} \label{sec:model_training}

\paragraph{BERT baseline} We build the outcome predictor by appending linear classification layer to the BERT embedding model,  with cross-entropy loss and sigmoid normalization for the binary classification task. We add a dropout layer with a rate of 0.3. Adam optimizer with an initial learning rate of 5e-5 is used, along with a scheduler
\(lr = lr_{init}/((1 +\alpha \cdot p)^\beta)\), where \(p = \frac{current\ epoch}{total\ epoch}\) and \(\alpha=10\) and \(\beta=0.25\), following~\cite{guo2023data}. 
To obtain the predicted outcome for all data, we randomly split the data into five folds. Each time we train the outcome predictor with four-fifths of the data, which is further split into training and validation sets by 80\%-20\% ratio. We train for 20 epochs and stop early when the model achieves the best validation F1 score. The batch size is 64. Finally, we predict on the remaining one-fifth data to obtain their predicted outcomes.

We use BERT model on huggingface (\url{https://huggingface.co/docs/transformers/en/model_doc/bert}) which is open sourced and permitted for research. The parameter size is about 110 millions.

\paragraph{\textsc{CausalDANN}}
Same to BERT baseline, we perform training and prediction by randomly splitting the data into five folds. We train with 20 epochs in total. However, during the first three epochs the model is trained without the domain adversarial module activated, the model to better learn the outcome labels first. The batch size is 32. The Adam optimizer, the learning rate and the scheduler are the same as those for the BERT baseline.

Equation~\ref{eq:DANN_loss} describes the loss to be minimized for DANN training. Following~\cite{guo2023data}, we balance the loss terms between the outcome predicting module and the domain predicting module by controlling $\lambda^{D}$ indirectly by
\begin{equation}
    \lambda^{D}=2/(1+e^{-\gamma \cdot p})-1
\end{equation}
where $p = \frac{current\ epoch - epochs\ w/o\ adv.}{total\ epochs}$, and $\gamma$ is now the hyperparameter. We performed a grid search for $\gamma$ in \([0.1, 1, 10]\) and decide to set $\gamma = 1$.

 We build our model on top of \url{https://github.com/NaJaeMin92/pytorch-DANN} and \url{https://github.com/fionasguo/DAMF} which are both open sourced and permitted for research. The parameter size is about 110 millions.

\paragraph{Inverse Propensity Weighting (IPW) Baseline}
The second baseline is Inverse Propensity Weighting (IPW), a popular causal estimation method. IPW accounts for confounders by adjusting an estimate with the inverse of the propensity score ${\pi(W,X)=P(T=1\mid W,X)}$. However, in our scenario, direct IPW is infeasible since intervened outcomes are unobserved. 

Nevertheless, IPW can also be seen as a sample reweighting technique, a commonly used domain adaptation technique~\cite{li2016prediction} that aligns the distributions of the observed non-intervened data and the unobserved intervened data. In our case, we define the interventions using an LLM or structured sampling (Section~\ref{sec:method_intervening}), and we can take advantage of the overlap assumption between the control and the intervened groups in the textual representation space. Thus, $\pi(W,X)$ is the probability of a piece of text being classified as interventional (as opposed to observational). This is equivalent to training a propensity model directly on observed binary treatments, to produce density ratios of interventional over observational. When propensity scores are accurately estimated, IPW helps debias outcome predictions. We can apply IPW as a baseline on top of predictions from vanilla BERT or \textsc{CausalDANN}.

We estimate the propensity score $\pi(W,X)$ using a separate BERT model, structured as in Section~\ref{sec:methods_BERT}. We train this propensity predictor on both intervened and non-intervened texts, using $T=\mathbbm{1}_{transformed}$ as labels. To obtain the propensity score as a probability, we calibrate it with temperature scaling~\cite{guo17} on a validation set that also includes intervened and non-intervened data.

Once the propensity scores are obtained, we compute the ATE as

{\small
\begin{align*}
    ATE &= \frac{1}{N}\sum_{i=1}^{N}\Big(\frac{\pi(W_i,X_i)}{1-\pi(W_i,X_i)}-1\Big) Y\numberthis
\end{align*}}


The CATE can be computed as

{\small
\begin{align*}
    CATE &= \frac{1}{N_{c,x}}\sum_{C_i=c,X_i=x}^{N_{1,c,x}}\Big(\frac{\pi(W_i,X_i)}{1-\pi(W_i,X_i)}-1\Big) Y \numberthis
\end{align*}}

The model archetecture, the training procedure and hyperparameters are the same as those for BERT baseline. The parameter size is about 110 millions.

\paragraph{Doubly Robust (DR) Baseline}
The third baseline is the doubly robust estimator~\cite{funk2011doubly}. DR combines two modeling approaches, the propensity score model and an outcome model, to estimate treatment effects. This ensures that the final estimation remains consistent if either the propensity score model or the outcome model is correctly specified. Further, the estimator's bias decays with the expected \emph{product} of the two models' errors, leading to double rate robustness. In our experiments, the ATE and CATE are computed as:

{\scriptsize
\begin{align*}
    ATE &= \frac{1}{N}\sum_{i=1}^{N}\Big[\Big(\frac{\pi(W_i,X_i)}{1-\pi(W_i,X_i)}-1\Big) (Y-\mu_{BERT}(W_i,X_i)) \\
    &+ \Big(\mu_{BERT}(g(W_i),X_i) - \mu_{BERT}(W_i,X_i)\Big)\Big] 
  \numberthis
\end{align*}}

{\scriptsize
\begin{align*}
    CATE &= \frac{1}{N_{c,x}}\sum_{C_i=c,X_i=x}^{N_{1,c,x}}\Big[\Big(\frac{\pi(W_i,X_i)}{1-\pi(W_i,X_i)}-1\Big) \\
    & \cdot (Y-\mu_{BERT}(W_i,X_i)) \\
    &+ \Big(\mu_{BERT}(g(W_i),X_i) - \mu_{BERT}(W_i,X_i)\Big)\Big]
 \numberthis
\end{align*}}

in which we use both propensity estimator $\pi(W_i,X_i)$ similar as in the IPW method and the outcome predictor $\mu_{BERT}(W_i,X_i)$ to ensure doubly robustness.

\paragraph{Computing resource} Each experiment was conducted on an RTX A6000 GPU with 48GB memory. On average, training and testing took 9.5 hours for BERT baselines and 11.2 hours for DANN models. The extended training time is due to the need for outcome and propensity prediction, each requiring five-fold cross-validation.

\section{Prompting LLMs} \label{sec:appendix_prompting}

\subsection{AITA-comments Data}
For the experiment AITA-comments in Section~\ref{sec:aita_comments}, we use the prompt in Table~\ref{code:aita_comments} to ask GPT-4 to provide a verdict based on a given AITA post and a top or a random comment. We remove the responses that GPT refuses to answer, possibly in cases where topics are sensitive.

\begin{table*}[h!]
\centering
\begin{tabular}{|p{15cm}|} 
\hline
\begin{small}
\begin{verbatim}
system_prompt = """\
  You are a moral arbiter hired by r/AmITheAsshole.\
  When presented with a story, you respond objectively with your verdict.\
  As is customary in that subreddit, you always begin your response with one of [NTA, YTA, ESH, NAH].
"""

def query_judge(story, comment):
  response = openai.ChatCompletion.create(
    model = model, temperature=1, max_tokens=2,
    messages = [
     {"role": "system", "content": system_prompt},
     {"role": "user", "content": f"The author said--\n{story}"},
     {"role": "user", "content": f"The community said--\n{comment}"},
     {"role": "user", "content": "What do you say? Is the verdict NTA, YTA, ESH, or NAH? No 
     explanations."}
    ])
  return response
\end{verbatim}
\end{small}\\
\hline
\end{tabular}
\caption{Code to prompt GPT-4 to generate verdicts based on a given AITA post and a comment.}
\label{code:aita_comments}
\end{table*}

The following Table~\ref{tab:aita_comment_generations} shows some example posts with their top and random comments, and the GPT generated verdicts. For some posts (e.g. row 1) the top and random comments are very similar in perspectives, and the GPT-generated verdicts also are the same when different comments are given. Interestingly,  the top and random comments in the second example have different perspectives. Nevertheless, GPT still gives the same verdict, possibly because the random comment was very short. The last example show how GPT changes verdict when the top and random comments are different in perspectives.

\begin{table*}[b]
    \centering
    \resizebox{1\linewidth}{!}{
    \begin{tabular}{p{500pt}|p{30pt}}
    \Xhline{1pt}
        Post and Comments & Verdict \\
        \hline
        \textbf{Post:} So I recently graduated high school and I'm attending a program lasting all of July that helps ease students into college life. My roomate met a girl there whom she really likes, and he told me and his other friends that he intends to ask her out to be his girlfriend. Keep in mind they just met. However, this girl told me that she does not reciprocate these feelings. I'm afraid to disclose this to my roommate, and I'm extremely confused if I'm the asshole in this situation. & \\
        \textbf{Top Comment:} I don't think you are an asshole for not telling him. It sounds like something he needs to hear from this girl. You can tell him if you feel it is the right thing to do. He may get angry with you and accuse you of having feeling for her. It can hurt not to have feeling reciprocated and sometimes & NAH \\
        \textbf{Random Comment:} You're not an asshole. Don't get involved. This is between your roommate and this girl, and it would be inappropriate and unwise to get between the two. & NAH \\
        \hline
        \textbf{Post:} My parents are diabetic, morbidly obese, and addicted to food. I'm the same, minus the diabetes. I've been trying for a while to eat healthily and lose weight. So have they, but yet every time they go to the store, they bring home junk food. I understand that it's hard for them, but it's also hard to stay on track with my healthier lifestyle when there's junk food all around me. Plus, they're getting older, so it is crucial for them to get healthy now. A few weeks ago, they brought home 4 bags of donuts, and I emptied each pack into the trash. Am I an asshole for this? **Edit:** I love all the responses. Got one "no", one "maybe", and one "yes". I agree with all of you. **Edit 2:** You guys are right--I'm the asshole. & \\
        \textbf{Top Comment:} Yep. You're the asshole. If you want to make the commitment to work hard on improving your health and your diet, then that is your job. If those who live with you want to support you by changing their lifestyle, too, that would be fabulous. But, it's not for you to dictate to them, or to throw away their property. & YTA \\
        \textbf{Random Comment:} What? No man. & YTA \\
        \hline
        \textbf{Post:} In a nutshell I was at a hardcore punk show. I got a beer between bands. I was drinking it up front waiting for the next band to start. They did and a ferocious pit opened up near me. So I got knocked around and so did my beer. I spilled over multiple people around me and they got beer over their clothes and hair. After I talked about this to some people who clearly don't go to punk shows they seemed to think I was rude, but frankly every person near the center of the crowd at such a show knows they'll possibly get beer spilled on them. It's happened to me multiple times. So AITA? & \\
        \textbf{Top Comment:} YTA. I go to local punk shows regularly and have worked a lot with booking agents and in music journalism. When the band takes the stage and the pit goes nuts you back to the side with the beer or put it down somewhere else. Spilling beer all over everyone is not normal, at least not in any of the & YTA \\
        \textbf{Random Comment:} Never been to a show like this. If it’s not common to take a drink where you know you’re going to get tossed around, YTA. If it’s basically expected to have a drink, then not the asshole. People can be annoyed they get beer spilled on them, but they know people have drinks and it can happen. Then they shouldn’t & NAH \\
    \Xhline{1pt}
    \end{tabular}}
    \caption{Example generations of verdicts by GPT-4 given AITA posts along with top or random comments.}
\label{tab:aita_comment_generations}
\end{table*}

\subsection{AITA-anger Data}
For the experiment AITA-anger in Section~\ref{sec:aita_anger}, we use the prompt in Table~\ref{code:aita_anger} to ask Claude-3.5-Sonet to (1) transform text to a version with higher anger, but keeping the semantic and all other linguistic properties the same; (2) rephrase text without changing the anger level (the purpose is to rephrase the non-intervened group too so that both groups are LLM-generated and bias from LLM generation can be mitigated), or (3) provide a verdict.

\begin{table*}[h!]
\centering
\begin{tabular}{|p{15cm}|} 
\hline
\begin{small}
\begin{verbatim}

def get_response(post, task):

    if task == 'anger_transformation':
        system_prompt = f"\
          Here is a Reddit post: {post}. Rephrase it while maintaining the semantics, writing style, 
          structures and paragraphs, but express much more anger. The output length should be 10 words 
          more or less than the original post. Please only output the rephrased texts:
        "
    elif task == 'rephrasing':
        system_prompt = f"\
          Here is a Reddit post: {post} Rephrase it while maintaining the semantics, writing style, 
          structures and paragraphs. The output length should be 10 words more or less than the
          original post. Please only output the rephrased texts:
        "
    elif task == 'give_verdict':
        system_prompt = f"\
          Given a Reddit post from r/AmItheAsshole, classify it into one of the following four 
          categories: \
          YTA (You're the asshole): The submitter is deemed to be in the wrong; \
          NTA (Not the asshole): The other party is considered to be wrong; \
          ESH (Everyone sucks here): Both parties are wrong; \
          NAH (No asshole): Both parties acted appropriately, but the situation just sucks. \
          Your task is to analyze the post and determine which category it belongs to based on the 
          narrative provided by the submitter. Consider the actions, intentions, and attitudes of all 
          involved parties to make an accurate classification. Only output one of these labels: YTA, 
          NTA, ESH, or NAH. No explanation is needed. Limit your response to 3 characters:
        "
    else:
        assert

    message = client.messages.create(
        model="claude-3-sonnet-20240229",
        max_tokens=1000,
        temperature=1,
        messages=[
          {
            "role": "user",
            "content": [
              {
                "type": "text",
                "text": system_prompt
              }]}]
    )
    return message.content[0].text
\end{verbatim}
\end{small}\\
\hline
\end{tabular}
\caption{Code to prompt Claude to transform a post to higher anger level but keeping other linguistic properties the same, or to rephrase a post, or to provide a verdict based on a given post.}
\label{code:aita_anger}
\end{table*}

We remove the responses that Claude refuses to answer, possibly in cases where topics are sensitive. The following Table~\ref{tab:aita_angry_generations} shows some examples of the rephrased and anger-transformed posts, as well as the Claude generated verdicts. We see that Claude has transformed the text into very aggressive and angry version. \textcolor{red}{Warning: Table~\ref{tab:aita_comment_generations} contains offensive language.}

\begin{table*}[b]
    \centering
    \resizebox{1\linewidth}{!}{
    \begin{tabular}{p{500pt}|p{30pt}}
    \Xhline{1pt}
        Posts & Verdict \\
        \hline
        \textbf{Original:} So I recently graduated high school and I'm attending a program lasting all of July that helps ease students into college life. My roomate met a girl there whom she really likes, and he told me and his other friends that he intends to ask her out to be his girlfriend. Keep in mind they just met. However, this girl told me that she does not reciprocate these feelings. I'm afraid to disclose this to my roommate, and I'm extremely confused if I'm the asshole in this situation. & \\
        \textbf{Rephrased:} Having recently completed high school, I'm participating in a July program designed to smoothly transition students into college life. My roommate encountered a girl there whom he's deeply attracted to, and he confided in me and our other friends his intention to ask her to be his romantic partner, despite their recent acquaintance. However, this young woman revealed to me that she doesn't share his amorous sentiments. I'm hesitant to divulge this information to my roommate, leaving me utterly perplexed about whether I'm the one in the wrong in this predicament. & NTA \\
        \textbf{Anger-Transformed:} I just finished that goddamn high school hell, and now I'm stuck in this idiotic college prep program for the entire fucking July. My moronic roommate met some chick there and instantly developed a pathetic crush on her, telling me and our other dumbass friends that he plans to ask her out as his girlfriend. Can you believe this imbecile's audacity? They barely know each other! But here's the real kicker: that girl confided in me that she doesn't share his delusional feelings at all. I'm torn between ripping the truth out of his thick skull or letting him make a fool of himself. Either way, I'm caught in this infuriating mess, and I can't decide if I'm the raging asshole here or not. & ESH \\
        \hline
        \textbf{Original:} My parents are diabetic, morbidly obese, and addicted to food. I'm the same, minus the diabetes. I've been trying for a while to eat healthily and lose weight. So have they, but yet every time they go to the store, they bring home junk food. I understand that it's hard for them, but it's also hard to stay on track with my healthier lifestyle when there's junk food all around me. Plus, they're getting older, so it is crucial for them to get healthy now. A few weeks ago, they brought home 4 bags of donuts, and I emptied each pack into the trash. Am I an asshole for this? **Edit:** I love all the responses. Got one "no", one "maybe", and one "yes". I agree with all of you. **Edit 2:** You guys are right--I'm the asshole. & \\
        \textbf{Rephrased:} My parents suffer from diabetes, extreme obesity, and food addiction. Like them, I'm obese but without diabetes. I've been striving to eat healthily and shed pounds for some time. They've also attempted this, yet they consistently bring unhealthy snacks home from the grocery store. While I understand their struggle, it's challenging to maintain my healthier lifestyle amidst the temptation of junk food. Moreover, as they age, it becomes crucial for them to prioritize their well-being. A few weeks ago, they purchased four bags of donuts, and I discarded the contents of each bag. Was I wrong for doing so? **Edit:** I appreciate all the responses – one "no," one "maybe," and one "yes." I concur with you all. **Edit 2:** You're correct – I was the one in the wrong. & ESH \\
        \textbf{Anger-Transformed:} My goddamn parents are fucking diabetic, disgustingly morbidly obese, and pathetically addicted to food like a bunch of pigs.  I'm the same, minus the shitty diabetes. I've been trying for a fucking eternity to eat healthily and lose weight, but those assholes keep sabotaging me. So have they, but yet every damn time they go to the store, they bring home junk food like the brainless morons they are. I understand that it's hard for those idiots, but it's also hard to stay on track with my healthier lifestyle when there's junk food all around me thanks to their stupidity. Plus, they're getting older, so it is crucial for them to get healthy now before they croak. A few weeks ago, those dumbasses brought home 4 bags of donuts, and I angrily emptied each pack into the trash like the responsible adult I am. Am I an asshole for this? **Edit:** I love all the responses, you fucking geniuses. Got one "no", one "maybe", and one "yes". I agree with all of you, you brilliant bastards. **Edit 2:** You guys are right--I'm the asshole, and I fucking hate myself for it. & ESH \\
    \Xhline{1pt}
    \end{tabular}}
    \caption{Example generations of rephrased and anger-transformed posts by Claude, and the verdicts also generated by Claude.}
\label{tab:aita_angry_generations}
\end{table*}

\section{BERTopic} \label{sec:appendix_bertopic}

To detect the top frequent topics in the AITA dataset, we use BERTopic. We first preprocess all posts to remove url links and stop words in English. We run BERTopic with UMAP dimension reduction and HDBSCAN clustering algorithms, and we use hyperparamters $top\_n\_words=10$, $n\_gram\_range=(1, 1)$, $min\_topic\_size=10$ and $nr\_topic=10$. Table~\ref{tab:bertopic} shows the top 30 topics detected and their post counts.

BERTopic assigns a single topic to each text. We focus on the top 30 topics, represented as a binary vector $\mathbf{C}$ for each post. The CATE is a vector of length equal to the number of topics, with each element representing the treatment effect averaged across data within that topic. In the IPW method we compute CATE similarly.

\begin{table*}[h]
    \centering
    \resizebox{0.4\linewidth}{!}{
    \begin{tabular}{l|c}
    \Xhline{1pt}
        Topic & Count \\
        \hline
        wedding\_birthday\_party\_family     & 204                       \\
        gift\_gifts\_christmas\_birthday     & 120                       \\
        job\_work\_boss\_for                 & 112                       \\
        teacher\_class\_school\_he           & 106                       \\
        racist\_white\_black\_english        & 104                       \\
        tip\_tipping\_food\_delivery         & 98                        \\
        ordered\_order\_pizza\_the           & 89                        \\
        friend\_him\_ex\_he                  & 80                        \\
        lane\_traffic\_road\_driving         & 66                        \\
        church\_religion\_religious\_atheist & 65                        \\
        dog\_dogs\_puppy\_him                & 64                        \\
        rent\_pay\_split\_bills              & 63                        \\
        her\_she\_girl\_me                   & 62                        \\
        child\_daughter\_have\_baby          & 60                        \\
        sex\_she\_porn\_relationship         & 56                        \\
        eat\_food\_cook\_she                 & 56                        \\
        name\_names\_my\_is                  & 54                        \\
        bathroom\_toilet\_shower\_use        & 52                        \\
        car\_drive\_gas\_pay                 & 50                        \\
        cat\_cats\_allergic\_my              & 49                        \\
        she\_go\_plans\_her                  & 47                        \\
        kids\_wife\_work\_babysit            & 47                        \\
        gay\_lgbt\_people\_men               & 42                        \\
        funeral\_died\_passed\_family        & 42                        \\
        he\_pair\_charger\_selling           & 39                        \\
        kid\_was\_the\_fish                  & 39                        \\
        hair\_shave\_beard\_cut              & 38                        \\
        wear\_wearing\_bra\_shirt            & 38                        \\
        eat\_food\_eating\_he                & 36                        \\
        tattoo\_art\_tattoos\_design         & 34                        \\
    \Xhline{1pt}
    \end{tabular}}
    \caption{Top 30 topics detected in AITA data by BERTopic.}
    \label{tab:bertopic}
\end{table*}

\section{Impact of Text-Length Filtering on AITA Results} \label{appendix:aita_text_length}
To reduce computational complexity, we select posts with fewer than 140 words. Our AITA semi-synthetic dataset contains approximately 12k samples after filtering out stories under 140 tokens. This is sufficient for training and testing. To ensure story length is not a major confounder, we conducted additional experiments on 10k randomly selected samples without length filtering. As shown in the table, CausalDANN still outperforms other baselines, confirming our conclusions.

\begin{table*}[h]
\centering
\resizebox{0.7\linewidth}{!}{
\begin{tabular}{l|cc|cc}
\Xhline{1pt}
\textbf{Model} & \multicolumn{2}{c|}{\textbf{AITA (filtered $<$140 tokens)}} & \multicolumn{2}{c}{\textbf{AITA (unfiltered)}} \\ \hline
              & $\delta$ATE & MSE of CATE & $\delta$ATE & MSE of CATE \\
\hline
BERT                  & 0.01 $\pm$ 0.01 & 0.01 $\pm$ 0.01 & 0.01 $\pm$ 0.01 & 0.01 $\pm$ 0.01 \\
IPW & $3.43 \pm 1.02$ & $14.08 \pm 9.03$ & $2.89 \pm 0.95$ & $11.46 \pm 6.23$ \\
DR            & 0.07 $\pm$ 0.03 & 0.04 $\pm$ 0.01 & 0.06 $\pm$ 0.01 & 0.05 $\pm$ 0.01 \\
\textsc{CausalDANN}   & 0.01 $\pm$ 0.01 & 0.00 $\pm$ 0.00 & 0.01 $\pm$ 0.01 & 0.01 $\pm$ 0.01 \\
\textsc{CausalDANN} + DR & 0.21 $\pm$ 0.06 & 0.08 $\pm$ 0.03 & 0.20 $\pm$ 0.07 & 0.09 $\pm$ 0.04 \\
\Xhline{1pt}
\end{tabular}}
\caption{Performance comparison on AITA data with and without story length filtering.}
\label{tab:aita_length}
\end{table*}

\end{document}